\documentclass[conference]{IEEEtran}
\IEEEoverridecommandlockouts
\usepackage{cite}
\usepackage{amsmath,amssymb,amsfonts}
\usepackage{algorithmic}
\usepackage{graphicx}
\usepackage{enumitem}
\usepackage{float}
\usepackage{booktabs}
\usepackage{textcomp}
\usepackage{multirow}
\usepackage{xcolor}
\usepackage{subcaption}
\usepackage{caption}

\usepackage{algorithm}
\usepackage{algorithmic}

\usepackage[
all=normal,
bibbreaks=tight,
paragraphs=tight,
floats=tight,
mathspacing=tight,
wordspacing=tight,
]{savetrees}

\usepackage{tabularx}
\newcolumntype{C}{>{\centering\arraybackslash}X}

\def\BibTeX{{\rm B\kern-.05em{\sc i\kern-.025em b}\kern-.08em
    T\kern-.1667em\lower.7ex\hbox{E}\kern-.125emX}}
\begin{document}

\title{HP-GMN: Graph Memory Networks for Heterophilous Graphs
}

\vskip -1em
\author{\IEEEauthorblockN{Junjie Xu, Enyan Dai, Xiang Zhang, Suhang Wang \linebreak}

\IEEEauthorblockA{The Pennsylvania State University, USA\\
\{junjiexu, emd5759, xzz89, szw494\}@psu.edu}}

% \and

\maketitle

\begin{abstract}
Graph neural networks (GNNs) have achieved great success in various graph problems. However, most GNNs are Message Passing Neural Networks (MPNNs) based on the homophily assumption, where nodes with the same label are connected in graphs. Real-world problems bring us heterophily problems, where nodes with different labels are connected in graphs. MPNNs fail to address the heterophily problem because they mix information from different distributions and are not good at capturing global patterns. Therefore, we investigate a novel Graph Memory Networks model on Heterophilous Graphs (HP-GMN) to the heterophily problem in this paper. In HP-GMN, local information and global patterns are learned by local statistics and the memory to facilitate the prediction. We further propose regularization terms to help the memory learn global information. We conduct extensive experiments to show that our method achieves state-of-the-art performance on both homophilous and heterophilous graphs. The code of this paper can be found at https://github.com/junjie-xu/HP-GMN. 

\end{abstract}


\vskip -1em
\section{Introduction} %1
% Graphs are pervasive in real-world, such as social networks \cite{wang2017signed, wu2020graph}, biological networks \cite{fout2017protein}, and information networks \cite{sun2013mining}. 
Graph Neural Networks have shown great ability in various graph tasks such as node classification \cite{kipf2017semi, dai2022comprehensive}, link prediction \cite{zhang2018link, liben2007link}, and graph classification \cite{zhang2018end}. Generally, the success of GNNs relies on the message-passing mechanism, where a node representation will be updated by aggregating the representations of its neighbors. Thus, the learned node representation captures both node attributes and local neighborhood information, which facilitates downstream tasks. The aggregation process of most current GNNs is explicitly or implicitly designed based on the homophily assumption \cite{zhu2020beyond}, i.e., two nodes of similar features or the same label are more likely to be linked. For example, in GCN, the representations are smoothed over connected neighbors with the assumption that the neighbors of a node are likely to have the same class and similar feature distribution. However, there are many heterophilous graphs in the real world that do not follow the homophily assumption. In heterophilous graphs, a node is also likely to connect to another node with dissimilar features or different labels. For example, fraudsters tend to contact normal users instead of other fraudsters in a trading network \cite{pandit2007netprobe}; different amino acids are connected to form functional proteins \cite{zhu2020beyond}; interdisciplinary researchers collaborate more with people from different areas in a citation network. 

Message Passing Neural Networks (MPNNs) with homophily assumption are challenged by heterophilous graphs. This is mainly because: (i) Since the majority of the neighborhood nodes lie in the same class as the center node in homophilous graphs, directly mixing the neighbor representations by averaging operation can preserve the context pattern, benefiting the downstream task. However, neighbors in heterophilous graphs come from different classes and have different distributions. The aggregation process in current MPNNs generally ignore the differences between nodes in different classes and simply mix them together. Therefore, the pattern of the local context in graphs with heterophily would not be well-preserved with the current traditional MPNNs; and (ii) only local context information of nodes is utilized in the prediction of MPNNs. The MPNNs generally fail to explicitly capture and utilize the global patterns of the nodes' local heterophilous context to give more accurate results.

\begin{figure}[t!]
\centerline{\includegraphics[width=0.9\linewidth]{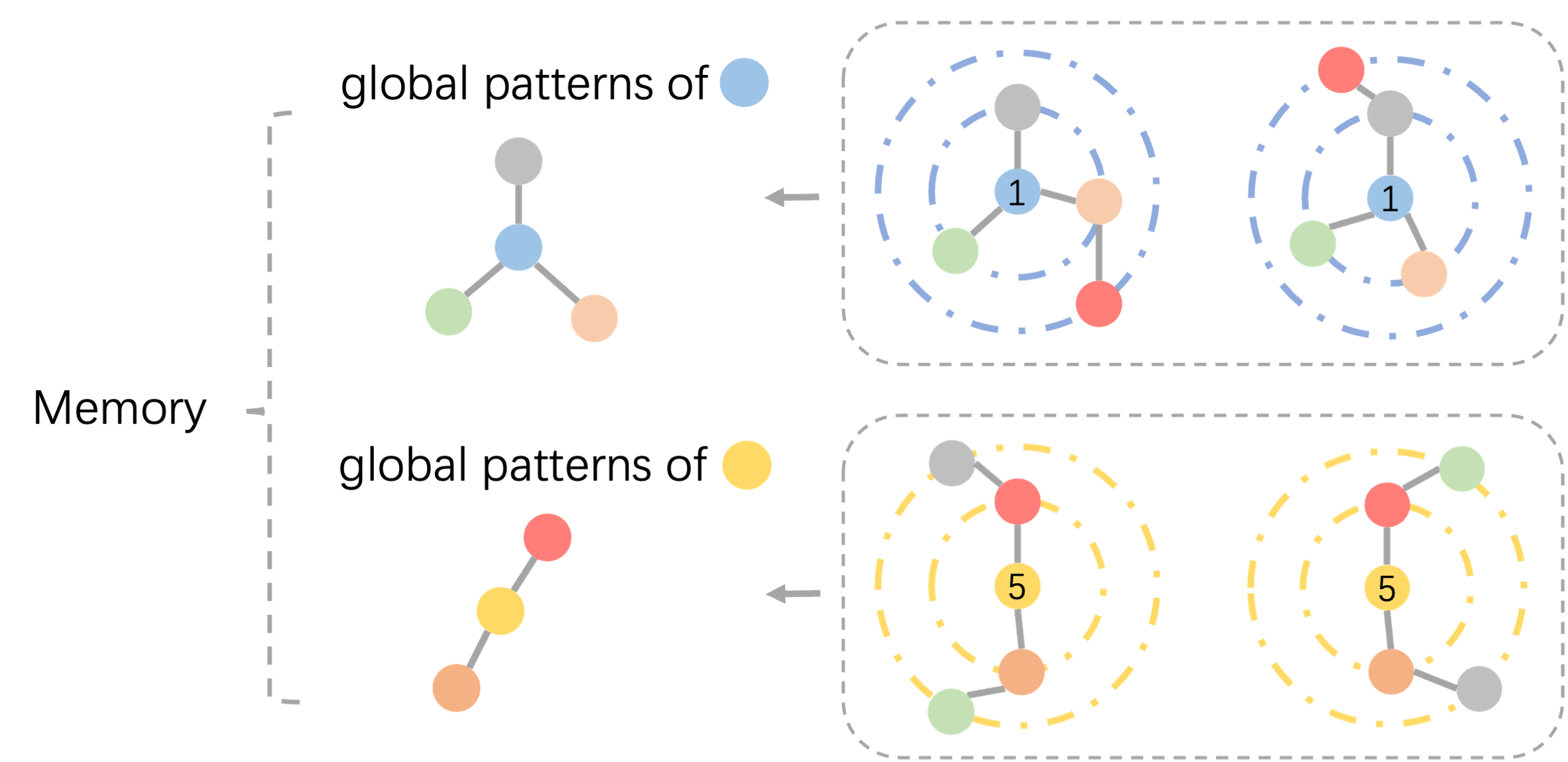}}
\vskip -0.3em
\captionsetup{font={scriptsize}}
\caption{Learning local statistics and global patterns on heterophilous graph. Color is the node label, while the number represents the attribute.}
\label{fig:intro}
\vskip -1em
\end{figure}

To overcome the shortcomings of MPNNs on heterophilous graphs, we propose the following two strategies. First, in heterophilous graphs,  though the neighbors of a node can have dissimilar node attributes and labels with the center node, we observe that nodes of the same class tend to have similar neighborhood distributions while nodes of different classes tend to have dissimilar neighborhoods. For example, as shown in the toy example in Fig. \ref{fig:intro}, nodes of ``blue" class have similar node attributes and subgraph structures, while nodes in the ``blue" class and the ``yellow" class have dissimilar attributes and subgraph structures. In addition, as Fig.~\ref{fig:intro} shows, neighbors' labels of blue and yellow nodes are different and blue nodes are connected to more neighbors. This indicates that the neighbors' distributions of ``blue" and ``yellow" are different. Thus, instead of simply aggregating the heterophilous neighbors' attributes which could result in noisy representations, we can summarize the local statistics from various aspects, such as features, the structure, and distributions of neighbors, to capture more comprehensive and discriminative information of the local context, which facilitates better representations. Second, each class has some representative and frequent subgraphs. Taking the ``blue" class as an example, we can observe that the blue nodes are generally connected with green, grey and orange nodes. The captured global patterns can benefit the prediction on the test instance by providing global information. However, current GNNs generally only rely on the representations of the local subgraph and fail to capture the global patterns of the nodes and their local context to facilitate the classification task. One promising method to address this issue is to learn global information by adopting Memory Networks \cite{ Weston2015MemoryN, tang2020joint}. Specifically, in memory networks, multiple memory units are used to store the global patterns of instances in different classes. The predictions can be given by matching the local patterns with the learned global patterns, which effectively utilize both the local context information and the global information. 

Therefore, this paper studies a novel problem of capturing both local and global information for heterophilous graphs. The main challenges are: \textbf{(1)} How to select local statistics that are good at capturing characteristics of subgraphs? \textbf{(2)} How to guarantee the memory stores global information favorable for the prediction? The memory units are required to be representative and diverse. Being representative makes sure that memory units record the most frequent patterns of the nodes while being diverse means that memory units are different from each other so that they will not record duplicate information. How should the update process be regulated to guarantee the diversity and representativeness of memory units? To deal with these issues, we propose \underline{G}raph \underline{M}emory \underline{N}etworks for \underline{H}etero\underline{p}hilous Graphs (\textbf{HP-GMN}). HP-GMN incorporates local statistics that can effectively capture the information of nodes in attributes, structures, and neighbor distributions on graphs with heterophily, and the memory that can learn global patterns of the graph. To ensure learning global patterns in high-quality, regularization methods are deployed to keep the memory units representative and diverse. 
% \suhang{you don't want to say various regularization terms, which makes your model sounds very complicated and ad hoc} 

In summary, we study the node classification task on heterophilous graphs, and the main contributions are:
\begin{itemize}[leftmargin=*]
    \item We develop a novel framework called HP-GMN using local statistics and the memory to learn local and global representations for heterophilous graphs.
    \item We propose regularization methods to encourage the update process of the memory to capture global information while keeping it diverse and representative.
    \item We conduct extensive experiments on real-world datasets and reveal our memory network outperforms state-of-the-art GNNs on both homophilous and heterophilous graphs.
\end{itemize}

\section{Preliminaries} %3
\label{sec:3_pre}

\subsection{Notations and Deﬁnition}
We use \(\mathcal{G} = (\mathcal{V}, \mathcal{E}, \mathbf{X})\) to denote an attributed graph, where \(\mathcal{V} = \{v_1,...,v_N\}\) is the set of $N$ nodes, and \(\mathcal{E} \subseteq \mathcal{V} \times \mathcal{V}\) is the set of edges.  \(\mathbf{X} = \{\mathbf{x}_1,...,\mathbf{x}_N\} \in \mathbb{R}^{|\mathcal{V}| \times F}\) is the node feature matrix, where \(\mathbf{x}_i\) is the node features of node \(v_i\) and $F$ is the feature dimension. \(\mathbf{A} \in \mathbb{R}^{N \times N}\) is the adjacency matrix, where \(\mathbf{A}_{ij}\) = 1 if nodes pair \(\{i, j\} \in \mathcal{E}\); otherwise \(\mathbf{A}_{ij}\) = 0. 

The graph neural networks aim to learn effective node representations \(\mathbf{H} \in \mathbb{R}^{|\mathcal{V}| \times F'}\) used in downstream tasks, where \(F'\) is the dimension of representations. Typically, GNNs use message passing \cite{gilmer2017neural} to learn representations. Each node aggregates its neighbors' representations and then combines them with the ego-representation. Hence, there are two essential steps in MPNNs:
\begin{equation}
\setlength{\abovedisplayskip}{4pt}
\setlength{\belowdisplayskip}{4pt}
\mathbf{a}_v^{k+1} = \text{AGGREGATE}^{k+1}(\{\mathbf{h}_u^k: u \in \mathcal{N}(v)\}) ,
\label{AGGREGATE}
\end{equation}
\begin{equation}
\setlength{\abovedisplayskip}{4pt}
\setlength{\belowdisplayskip}{4pt}
\mathbf{H}_v^{k+1} = \text{COMBINE}^{k+1}(\mathbf{h}_v^k, \mathbf{a}_v^k) , \label{COMBINE}
\end{equation}
where $\mathbf{h}_v^k$ is the representation of node $v$ at layer $k$ and $\mathcal{N}(v)$ is the neighbors of node $v$. AGGREGATE and COMBINE are two functions specified by GNNs.

Generally, the homophily degree of a graph can be measured by node homophily ratio \cite{Pei2020Geom-GCN:}. 

\textbf{Definition1 (Node homophily ratio)} \cite{Pei2020Geom-GCN:} \textit{It is the average ratio of same-class neighbor nodes to the total neighbor nodes in a graph.}% Edge homophily ratio measures the ratio of edges connecting nodes with the same label to all the edges.}
\begin{equation}
\setlength{\abovedisplayskip}{4pt}
\setlength{\belowdisplayskip}{4pt}
\mathcal{H}_{node} = \frac{1}{|V|}\sum_{v \in \mathcal{V}} \frac{|\{u \in \mathcal{N}(v): y_v = y_u\}|}{|\mathcal{N}(v)|} \in [0, 1] \ ,
\end{equation}
% \begin{equation}
% \mathcal{H}_{edge} = \frac{|\{(v, u) \in \mathcal{E}: y_v = y_u\}|}{|\mathcal{E}|} ,
% \end{equation}
where $y$ is the node label. Graphs with higher homophily are close to 1, and graphs with higher heterophily are close to 0.

\subsection{Intuition and Problem Definition}
There are two main reasons leading to the incompatibility between MPNNs and heterophily. Firstly, AGGREGATE is usually implemented by a mean or weighted mean function and COMBINE is to mix a node's ego- and neighbor- representations. Because neighbors are from different classes and have different distributions in the heterophily assumption, such process can mix representations from different distributions and make the result of aggregation less discriminative. This mixture gives us poor node representations, which further degrades the quality of global patterns learned from each node. In this case, even MLP can have better performance than GCN because the structure information is harmful according to some empirical results \cite{zhu2020beyond, ma2022is}. Secondly, in MPNNs, a node only gathers information from the local neighbors and fails to get more global information. As a result, MPNNs cannot learn global patterns from nodes' local contexts. Therefore, MPNNs are not good at addressing the heterophily problem. 

In a graph, each node has a $k$-hop subgraph centered on itself. The characteristics of the subgraph can provide much information for the node's local context; therefore, local statistics can be designed to capture this local information for downstream classification tasks. Furthermore, for all the nodes of each class, we can learn the global patterns of the class by extracting representative information of the nodes' local representations. Then each node can learn global information from similar global patterns. We need to measure the similarity here because we do not need information from different distributions. Based on the high-level ideas, we develop HP-GMN to learn from both local and global information. To measure the performance of our framework, we mainly have semi-supervised node classification as the downstream task. 

\textbf{Problem1} \textit{Given an attributed graph \(\mathcal{G} = (\mathcal{V}, \mathcal{E}, \mathbf{X})\) and a training set \(\mathcal{V_L}\ \subseteq \mathcal{V}\), with known class labels \(\mathcal{Y}_L= \{y_v, \forall v \subseteq \mathcal{V_L} \} \), the heterophily ratio of the graph $\mathcal{G}$ is low, we aim to learn a GNN that can infer the unknown class labels \(\hat{\mathcal{Y}}_U = \{\hat{y}_u, \forall u \subseteq \mathcal{V_U} = (\mathcal{V} - \mathcal{V_L}) \}\). }

\begin{figure}[t!]
\centerline{\includegraphics[width=0.98\linewidth]{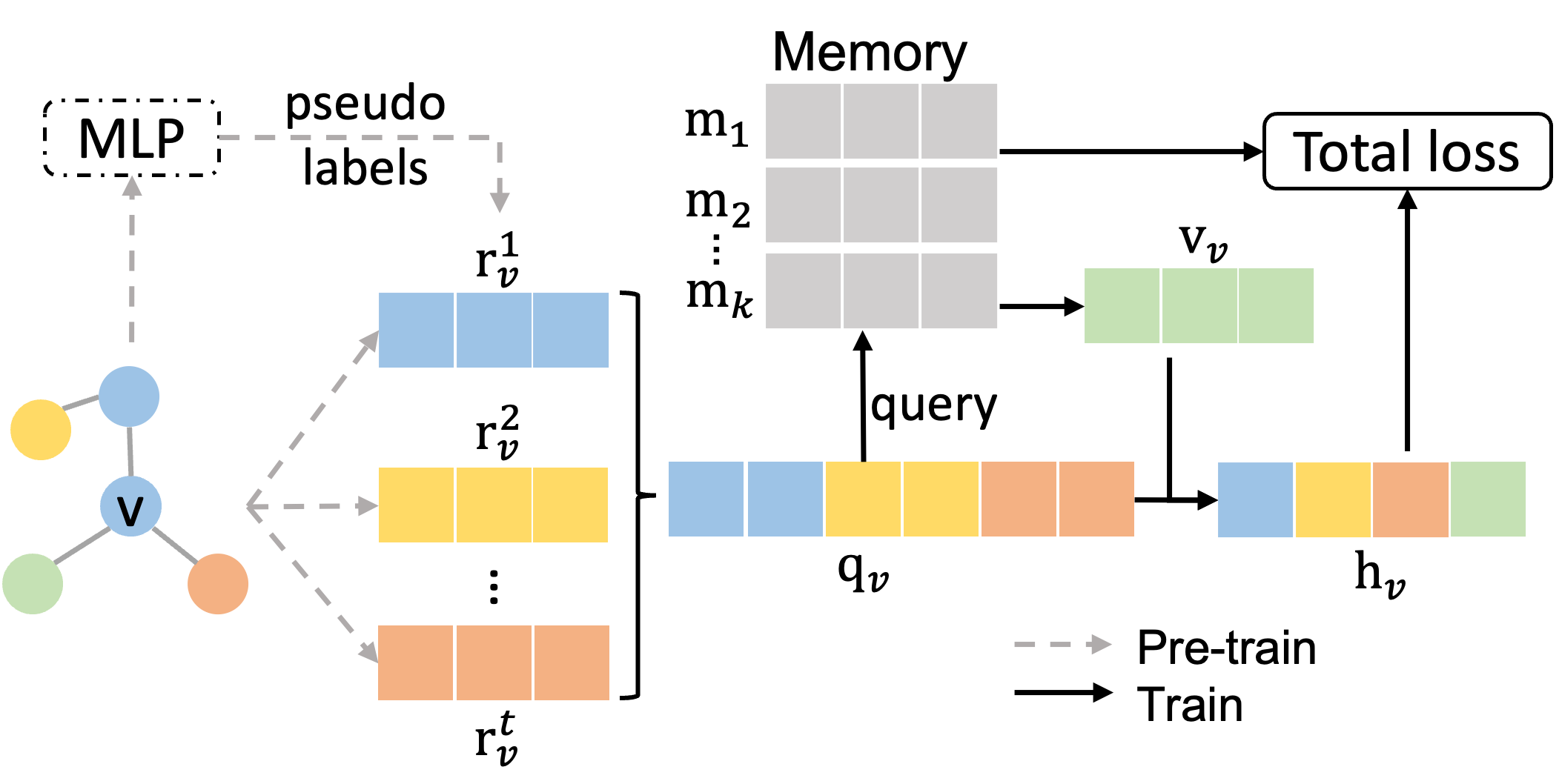}}
\captionsetup{font={scriptsize}}
\caption{The framework of HP-GMN. Informative local statistics \{$\mathbf{r}^1, \cdots, \mathbf{r}^t$\} are summarized and transformed to obtain local pattern vectors $\mathbf{q}$. As for the memory network module, it deploys a memory matrix $\mathbf{M}$ consisting of memory units to store global patterns of the nodes in different classes. $\mathbf{q}$ will be used to query the memory matrix to give the value vector $\mathbf{v}$.  }
\vskip -1em
\label{fig:framework}
\end{figure}

\section{Methodology} %4

% \enyan{The idea is not clearly described here.}
In this section, we describe our proposed HP-GMN thoroughly. The overall idea is to utilize local statistics to capture useful local information and simultaneously deploy the memory network to extract global patterns. Then, accurate predictions can be given by combining both their local context information and the captured global patterns. There are several challenges in the process: (1) How to design local statistics that can effectively describe the local context of nodes in heterophilous graphs? (2) How to obtain representative and diverse memory units to capture informative global patterns? In an attempt to address these challenges, node attributes and diffusion matrix are incorporated into the local statistics to preserve the local pattern in attribute and structure information. In addition, label-wise statistics~\cite{dai2021labelwise} are also used to describe neighbors' distributions of center nodes. To ensure the quality of the memory units, Kpattern regularization and Entropy regularization are adopted to guarantee diversity and representativeness. The overall framework of the proposed HP-GNN is shown in Fig.~\ref{fig:framework}. 
% Specifically,  In the query process, we calculate attention scores between $\mathbf{q}_v$ and each memory unit $[\mathbf{m}_1, \mathbf{m}_2, \cdots, \mathbf{m}_k]^\mathrm{T}$. We aggregate global information from these memory units to form the global representation $\mathbf{v}_v$ of node $v$ according to the attention scores. Finally, the local and global representations are concatenated to form the final node representation. 
% Next, we introduce each part in details.

% \suhang{don't simply describe the framework, connect back to the challenges and how we design the framework to address the challenges}

% In section 1, we introduce how to select the local statistics. In section 2, we clarify how to get the query and value of the memory. In section , we discuss several potential loss functions to regularize the readout and update behaviors of the memory. 

\subsection{Local Representation with Local Statistics} \label{local_stat}
It has been observed that the local context of nodes in heterophilous graphs can differ a lot for nodes are in different classes, while two nodes in the same class often exhibit similar node features and local subgraphs~\cite{dai2021labelwise}. Thus, we propose the local statistics including node attributes, neighborhood distributions, and local topology.

\subsubsection{Node Attributes} 
As node attributes contains important information about node labels and similar nodes tend to have similar attributes, the first local statistic is given by $\mathbf{r}^1_v = \mathbf{x}_v$, where $\mathbf{x}_v$ denotes the attributes of node $v$.

\subsubsection{Label-wise Statistics} 
To effectively preserve the neighborhood distribution, we adopt the label-wise aggregation, which has been proved to be an effective method for heterophilous graphs \cite{dai2021labelwise} because it aggregates neighbors from different classes separately and avoids obscuring the boundaries between classes. In this paper, we use two label-wise characteristics as local statistics, i.e., label-wise neighbor class distribution and label-wise neighbor feature distribution.

First, as suggested by \cite{dai2021labelwise}, we estimate pseudo labels of the unlabelled nodes, which are used to guide the label-wise aggregation. Speciﬁcally, a label estimator $f_E$ is utilized to obtain the pseudo label of node $v$: $\hat{y}_v^{\ pseudo} = f_E(x_v)$.
Many classifiers can be used as the estimator $f_E$, such as GCN or MLP. We adopt MLP in our framework for simplicity. $f_E$ is trained with labelled nodes.

We assume similar nodes to have similar local neighbor class distributions. Thus, for each node $v$, label-wise neighbor class distribution counts the numbers of neighbors of each class using the pseudo labels:
\begin{equation}
\setlength{\abovedisplayskip}{4pt}
\setlength{\belowdisplayskip}{4pt}
\mathbf{r}^2_v = \big[ \ \big|\mathcal{N}_1(v) \big| \ , \ \big|\mathcal{N}_2(v) \big|, \cdots, \big|\mathcal{N}_{|\mathcal{Y}|}(v) \big| \ \big] ,
\end{equation}
where $\mathcal{N}_i(v) = \{u:(v,u) \in  \mathcal{E}\ and\ \hat{y}_u = i \}$ denote the set of neighbors of $v$ with pseudo label $i$ and $|\cdot|$ is the size of a set. %Note that the neighbors here is the label-wise neighbors, i.e. neighbors with the same label $\mathcal{N}_i(v) = \{u:(v,u) \in  \mathcal{E}\ and\ \hat{y}_u = i \}$, and 
Hence, $|\mathcal{N}_i(v)|$ gives the number of neighbors of node $v$ with label $i$.

Similarly, we expect similar nodes in heterophilous graphs to have similar local neighborhood feature distributions. Thus, for each node, label-wise neighbor feature distribution calculates the average features of neighbors from each label as:
\begin{equation}
\setlength{\abovedisplayskip}{4pt}
\setlength{\belowdisplayskip}{4pt}
\mathbf{r}^3_v = \Big[\sum_{u \in \mathcal{N}_1(v)} \frac{\mathbf{x}_u}{|\mathcal{N}_1(v)|} \ \Big|\Big| \ \cdots \  \Big|\Big| \sum_{u \in \mathcal{N}_{|\mathcal{Y}|}(v)} \frac{\mathbf{x}_u}{|\mathcal{N}_{|\mathcal{Y}|}(v)|} \ \Big] ,
\end{equation}
% \begin{equation}
% \mathbf{r}^3_v \leftarrow \operatorname{MLP} (\mathbf{r}^3_v).
% \end{equation}
where $\|$ denotes the concatenation operation. 
% \suhang{could you explain why we add MLP?}

\subsubsection{Diffusion Matrix}
Structure information is essential and similar nodes often have similar structures in their subgraphs. We exploit higher-order structural information by utilizing the diffusion matrix. The diffusion matrix can remove the restriction of using only the direct 1-hop neighbors through different powers of the adjacency matrix \cite{klicpera2019diffusion}. It is as:
\begin{equation}
\setlength{\abovedisplayskip}{4pt}
\setlength{\belowdisplayskip}{4pt}
\mathbf{S}=\sum_{k=0}^{\infty} \theta_{k} \mathbf{T}^{k}  ,
\end{equation}
where $\mathbf{s}_v$ is the $v$-th row of matrix $\mathbf{S}$ and $\mathbf{T}$ is the transition matrix. We require that $\sum_{k=0}^{\infty} \theta_{k}=1, \theta_{k} \in[0,1]$, and the eigenvalues of $\mathbf{T} \in[0,1]$ to guarantee the convergence. The transition matrix $\mathbf{T}$ and weighting coefficients $\theta_{k}$ are defined in various ways. In this paper, we use the PPR diffusion matrix \cite{page1999pagerank} due to its good performance and flexibility:
\begin{equation}
\setlength{\abovedisplayskip}{4pt}
\setlength{\belowdisplayskip}{4pt}
\mathbf{T}=\mathbf{A} \mathbf{D}^{-1} ,\ \theta_{k} = \alpha(1-\alpha)^{k} ,
\end{equation}
where teleport probability $\alpha \in (0, 1)$. 
% The graph diffusion with heat kernel is given by:
% \begin{equation}
% \mathbf{T}=\mathbf{A} \mathbf{D}^{-1} ,\ \theta_{k} = e^{-t} \frac{t^{k}}{k!} ,
% \end{equation}
% where $t$ is the diffusion time. We use the PPR diffusion matrix in our model. 

\subsubsection{Local Representation} With the above four types of features capturing different perspectives of the local statistics, we transform them by MLPs since some local statistics might be high dimensional, and transformation to the latent space can reduce the dimension and better reflect their patterns. Finally, we concatenate them to form the local representation $\mathbf{q}_v$ of node $v$ as: %\suhang{in the following, please call it as local presentation instead of local statistics}
\begin{equation}
\setlength{\abovedisplayskip}{4pt}
\setlength{\belowdisplayskip}{4pt}
\mathbf{q}_v = [\operatorname{MLP}_1(\mathbf{r}^1_v) || \operatorname{MLP}_2(\mathbf{r}^2_v) || \operatorname{MLP}_3(\mathbf{r}^3_v) || \operatorname{MLP}_4(\mathbf{r}^4_v)].
\end{equation}

In summary, the advantages of $\mathbf{q}_v$ are: (i) it avoids aggregating neighbors of dissimilar features and captures node attributes and heterophilous local graph information. (ii) it can be used to query the memory to get better global information.

\subsection{Global Representation Learning with Memory Unit} %4.2
The global patterns are some representative local graph patterns that appear most frequently for nodes of each class. We want to learn them so that nodes can aggregate from similar patterns instead of dissimilar neighbors. However, directly learning representative subgraphs is difficult as we need to consider both graph structure and node attributes. Thus, we learn representative global vector representations using a memory module instead. Specifically, the memory is a matrix $\mathbf{M} \in \mathbb{R}^{K \times hidden} = [\mathbf{m}_1, \mathbf{m}_2, \cdots, \mathbf{m}_K]^\mathrm{T}$, where $K$ is the number of memory units and $hidden$ is the feature dimension of a memory unit. Each row $\mathbf{m_i}$ is a memory unit used to learn a global pattern. 

To get the global representations, we use the attention scores to measure the similarity between each node's local representation and every global pattern. Then a node gathers information from each pattern according to the attention scores. Intuitively, the node get global information by the combinations of the patterns that can describe the distribution it belongs to. We follow \cite{sukhbaatar2015end} to query the memory and get attention matrix:
\begin{equation}
\setlength{\abovedisplayskip}{4pt}
\setlength{\belowdisplayskip}{4pt}
\mathbf{S} = \operatorname{Softmax}(\mathbf{M}\mathbf{Q}^\mathrm{T}) ,
\end{equation}
where $\mathbf{Q} = [\mathbf{q}_1, \mathbf{q}_2, \cdots, \mathbf{q}_n]^\mathrm{T}$ 
% \suhang{we have $N$ queries, not $k$. You really need to pay attention to your equations.} 
is the query matrix. $s_{ij}$ indicates the importance of the $i$-th memory to $j$-th node. Then we calculate the value matrix \(\mathbf{V}\), which is global representation by weighted average of the memories as:
\begin{equation}
\setlength{\abovedisplayskip}{4pt}
\setlength{\belowdisplayskip}{4pt}
\mathbf{V} = \mathbf{S}^\mathrm{T} \mathbf{M} .
\end{equation}

Eventually, we concatenate the local and global representation of a node followed by an MLP transformation to get the final representation $\mathbf{h}_v$. Then we can predict the label distribution of $\hat{\mathbf{y}}_v$.
\begin{equation}
\setlength{\abovedisplayskip}{4pt}
\setlength{\belowdisplayskip}{4pt}
\label{eq_finalrep}
\mathbf{h}_v = \operatorname{MLP}_5([\mathbf{q}_v||\mathbf{v}_v]) ,
\end{equation}
\begin{equation}
\setlength{\abovedisplayskip}{4pt}
\setlength{\belowdisplayskip}{4pt}
\hat{\mathbf{y}}_v = \operatorname{Softmax}(\mathbf{h}_v) ,
\end{equation}
where $\hat{\mathbf{y}}_i$ is the predicted label probabilities of node $v_i$. 

For the training, we minimize the cross-entropy loss as
\begin{equation}
\setlength{\abovedisplayskip}{4pt}
\setlength{\belowdisplayskip}{4pt}
\mathcal{L}_{class} = \frac{1}{|\mathcal{V}_{train}|} \sum_{v \in \mathcal{V}_{train}} l(\hat{\mathbf{y}}_{v}, \mathbf{y}_{v}) ,
\end{equation}
where $\mathcal{V}_{train}$ is the set of labeled nodes, $\mathbf{y}_{v}$ is the one-hot-encoding of $v$'s label and  $l(\cdot, \cdot)$ is the cross entropy loss.

\subsection{Memory Regularization} %4.2.3
In order to capture global information effectively, we desire two properties of the memory units, i.e., representativeness and diversity. Representativeness is desired because the memory is expected to capture the most representative patterns of the distributions of a class. Diversity is also required to encourage richer information and avoid much-duplicated information recorded in memory units. However, during the training process, memory units are updated by gradient descent. The learned memory units are likely to lack the properties we desire. Therefore, we propose Kpattern and Entropy regularizations to encourage the properties.

\begin{table}[t!]
\captionsetup{font={scriptsize}}
\caption{Statistics of heterophilous and homophilous datasets.}
\vspace{-0.5em}
\begin{tabular}{ccccccc}
\toprule
\textbf{Datasets}  & \textbf{\#Nodes} & \textbf{\#Edges} & \textbf{\#Attributes} & \textbf{\#Classes} & {$\mathcal{H}_{node}$} \\ \hline
\textbf{Texas}     & 183              & 309              & 1,703        & 5            & 0.06                       \\
\textbf{Wisconsin} & 251              & 499              & 1,703        & 5            & 0.16                        \\
\textbf{Cornell}   & 183              & 295              & 1,703        & 5            & 0.11                       \\ \hline
\textbf{Chameleon} & 2,277            & 36,101           & 2,325        & 5            & 0.25                        \\
\textbf{Squirrel}  & 5,201            & 217,073          & 2,089        & 5            & 0.22                       \\
\textbf{Crocodile} & 11,631           & 360,040          & 128          & 5            & 0.30                       \\ \hline
\textbf{Cora}      & 2708             & 5429             & 1433         & 7            & 0.83                      \\
\textbf{Citeseer}  & 3327             & 4732             & 3703         & 6            & 0.71                       \\
\textbf{Pubmed}    & 19717            & 44338            & 500          & 3            & 0.79                    \\ 

\bottomrule
\label{tab:Statistics}
\vspace{-3em}
\end{tabular}
\end{table}

% \begin{table*}[t!]
% \caption{Statistics of Heterophilous and Homophilous Graph Datasets.}
% % \suhang{this takes too much space, the original one column table is good, you can save space for algorithm if you have time to revise}}
% \vspace{-1em}
% \begin{center}
% \begin{tabularx}{0.85\linewidth}{p{0.08\linewidth}|CCC|CCC|CCC}
% \toprule
% \multirow{2}{*}{\textbf{Dataset}} &
%   \multicolumn{3}{c|}{\textbf{WebKB}} &
%   \multicolumn{3}{c|}{\textbf{Wikipedia}} &
%   \multicolumn{3}{c}{\textbf{Citation}} \\ &
%   \textbf{Texas} &
%   \textbf{Wisconsin} &
%   {\textbf{Cornell}} &
%   \textbf{Chameleon} &
%   \textbf{Squirrel} &
%   {\textbf{Crocodile}} &
%   \textbf{Cora} &
%   \textbf{Citeseer} &
%   \textbf{Pubmed} \\ \midrule
% {Nodes}                          &183    &251    &183    &2277   &5201       &11,631     &2708  &3327  &19717    \\
% {Edges}                          &309    &499    &295    &36101  &217,073    &360040     &5429  &4732  &44338    \\
% {Attributes}                     &1703   &1703   &1703   &2325   &2089       &128        &1433  &3703  &500      \\
% {Classes}                        &5      &5      &5      &5      &5          &5          &7     &6     &3        \\
% {$\mathcal{H}_{node}$}           &0.06   &0.16   &0.11   &0.25   &0.22       &0.30       &0.83  &0.71  &0.79     \\
% {$\mathcal{H}_{edge}$}           &0.11   &0.20   &0.31   &0.24   &0.22       &0.25       &0.81  &0.74  &0.80     \\

% \bottomrule
% \end{tabularx}
% \end{center}
% \label{tab:Statistics}
% \vspace{-1.5em}
% \end{table*}

\textbf{Kpattern Regularization}. 
To make the memory representative, we want each query $\mathbf{q}_v$ to be at least close to one of the memory units so that each query can retrieve important global information. In other words, the memory should be useful for each node. Thus, we calculate the distances between each node's local representation and every memory unit. Then we choose the memory unit with the smallest distance. We sum up all the smallest distances over all nodes, and we aim to learn memory that can minimize the total loss as:
\begin{equation}
\label{eq_kpattern}
\setlength{\abovedisplayskip}{4pt}
\setlength{\belowdisplayskip}{4pt}
\min_{\mathbf{M}} \mathcal{L}_{kpattern} = \min_{\mathbf{M} }\sum_{v \in \mathcal{V}} \min_{\mathbf{m}_i \in \mathbf{M}} dist(\mathbf{m}_i, \mathbf{q}_v),
\end{equation}
where $dist$ can be any function measuring distance between two vectors. We adopt the Euclidean distance in this paper. Intuitively, we prefer representative $\mathbf{M}$ that makes the total distance between the memory and the query minimized, which means the $K$ patterns become representative for the query.

\textbf{Entropy Regularization}. 
To make the memory diverse, we want each memory to have similar chances of being selected globally. In other words, the averaged importance score over all the nodes for each memory should be similar to each other. Since $s_{ij}$ indicates the importance of $i$-th memory to $j$-th node, the overall importance score of $\mathbf{m}_i$ can be written as $\mathbf{s}_i' = \sum_{j=1}^{N} s_{ij}$.
Denote the memory importance vector as $\mathbf{s}' \in \mathbb{R}^{K}$ with $\mathbf{s}_i'$ as the $i$-th element of $\mathbf{s}'$ indicating the total importance of $i$-th memory unit. 

We need the elements in $\mathbf{s}'$ to be equal or similar to others, i.e., $\mathbf{s}'$ is uniformly distributed, which means all the memory units are equally important and are used with the same frequency. Otherwise, if some units have low overall importance scores than others, there are two possible reasons. (1) Some memory units are unimportant because they contain duplicated information from others. As a result, the information contained in these units can also be expressed by others. (2) Some memory units are useless, i.e., not representative. They do not contain any useful information demanded by downstream tasks, so they are rarely used. 
Therefore, we can achieve uniformly distributed $\mathbf{s}'$ by maximizing the entropy of it, namely, i.e., minimizing the negative entropy as:
\begin{equation}
\setlength{\abovedisplayskip}{4pt}
\setlength{\belowdisplayskip}{4pt}
\label{eq_entropy}
\min_{\mathbf{M}} \mathcal{L}_{entropy} = - \mathrm{H}(\mathbf{s}') 
\end{equation}

\subsection{Final Objective Function of HP-GMN}
With $\mathbf{h}_i$ in Eq.(\ref{eq_finalrep}) capturing both local and global information, $\mathcal{L}_{kpattern}$ in Eq.(\ref{eq_kpattern}) to make the memory representative and $\mathcal{L}_{entropy}$ in Eq.(\ref{eq_entropy}) to make the memory diverse, the final objective function of HP-GMN is:
\begin{equation}
\setlength{\abovedisplayskip}{4pt}
\setlength{\belowdisplayskip}{4pt}
\min_{\theta,\mathbf{M}}\mathcal{L}_{total} = \mathcal{L}_{class} + \alpha\mathcal{L}_{kpattern} + \beta\mathcal{L}_{entropy} ,
\end{equation}
where $\alpha$ and $\beta$ are scalars to control the contributions of $\mathcal{L}_{kpattern}$ and $\mathcal{L}_{entropy}$. $\theta$ is the set of parameters of MLPs.

% \suhang{when you write down loss function, add minimize or maximize with respect to the parameters.} \suhang{discuss a little bit about how we train the model. e.g., use gradient descent or something. You can add the training algorithm if you have time and brief explain the algorithm here}

% \begin{algorithm}[t] 
%     \caption{ Training Algorithm of .} 
%     \begin{algorithmic}[1]
%         % \REQUIRE $\mathcal{D}$, $K$, $M$, $T_l$, $T_h$, $\alpha$, $\beta$, $\tau$
%         % \ENSURE $f_G$, $f_E$, $f_C$, and $\mathcal{\Tilde H}$.

%         \STATE Pretrain the $f_G$ and $f_E$ with Eq.(\ref{eq:pretrain})
%         \STATE Obtain the initialized prototypes and embeddings with Eq.(\ref{eq:get_init})
%         \REPEAT
%         \STATE Feed $\mathcal{\Tilde H}$ into the $f_G$ to generate the attributes and structure of  prototype graphs with $f_G$ by Eq.(\ref{}) and Eq.(\ref{})
%         \STATE Jointly optimize prototype embeddings  $\mathcal{\Tilde H}$ and the parameters of $f_G$ $f_E$, and $f_C$ by Eq.(\ref{})
%         \UNTIL convergence
%         \RETURN $f_G$, $f_E$, $f_C$, and $\mathcal{\Tilde H}$.
%     \end{algorithmic}
%     \label{alg:1}
% \end{algorithm}

\section{Experiments} %5
In this section, we conduct extensive experiments to demonstrate the effectiveness of our proposed framework. In particular, we aim to answer the following research questions: \textbf{RQ1} How effective is the Memory Network in the node classification on heterophilous graphs? \textbf{RQ2} Can HP-GMN be extended to homophilous graphs? \textbf{RQ3} What are the contributions of each component to the final results?

\subsection{Experimental Settings} %5.1
\subsubsection{Baselines} We compare our method with representative GNNs (MLP, GCN), GNNs incorporating high-order neighborhood information (MixHop \cite{abu2019mixhop}, GCN+Jump Knowledge \cite{xu2018representation}, APPNP \cite{klicpera2018predict}), as well as GNNs for heterophily (BMGCN \cite{he2021block}, GPRGNN \cite{chien2021adaptive}, MMP \cite{chen2022memory}).

\begin{table}[t!]
\captionsetup{font={scriptsize}}
\caption{Node classiﬁcation performance on heterophily. (Accuracy(\%) ± Std.)}
\vspace{-0.7em}
\begin{center}
\begin{tabularx}{1.0\linewidth}{XCCCCCC}
\toprule

Methods & \textbf{Texas} & \textbf{Wisc.} & \textbf{Cornell} & \textbf{Cham.} & \textbf{Squi.} & \textbf{Croco.} \\
\midrule
MLP         &80.8±7.0       &85.1±5.6         &83.2±6.4      &48.3±2.2          &33.4±1.1          &65.7±1.0            \\
GCN         &59.5±7.1       &51.4±3.1         &56.8±4.5      &67.5±2.6          &56.6±1.4          &71.9±0.9            \\
MixHop      &78.7±6.3       &83.7±4.6         &79.5±6.3      &49.3±1.2          &49.3±1.2          &73.9±1.2            \\
GCN+JK      &61.6±7.2       &58.4±4.1         &59.2±7.0      &52.6±1.3          &52.6±1.3          &71.9±0.8            \\
APPNP       &80.5±3.8       &84.7±3.9         &83.0±7.2      &40.4±1.6          &40.4±1.6          &66.3±1.1            \\
GPRGNN      &81.4±5.2       &82.8±3.8         &77.6±7.1      &69.3±1.4          &49.6±1.7          &68.2±0.8            \\
BMGCN       &80.0±5.4       &75.3±6.1         &74.6±5.0      &69.0±1.6          &52.7±1.1          &64.3±1.1            \\
MMP         &77.6±6.0       &84.1±4.0         &79.5±8.5      &66.1±2.2          &53.1±10.9         &66.2±0.4            \\
\textbf{HPGMN}  &\textbf{85.1±4.2}       &\textbf{86.5±3.1}         &\textbf{84.1±5.3}      &\textbf{79.6±0.8}          &\textbf{72.3±2.3}          &\textbf{80.8±0.3}            \\

\bottomrule
\end{tabularx}
\end{center}
\label{tab:results}
\vspace{-2em}
\end{table}

\subsubsection{Datasets} We utilize six publicly available heterophilous datasets, including three WebKB datasets (Cornell, Texas, Wisconsin), and three Wikipedia network datasets (Chameleon, Squirrel, Crocodile) \cite{Pei2020Geom-GCN:}. We use three homophilous datasets from citation networks \cite{sen2008collective} (Cora, Citeseer, PubMed). The key statistics of the datasets are summarized in Table \ref{tab:Statistics}.

% \textbf{WebKB}: WebKB is a collection of webpages from CS departments of several universities. It was processed by \cite{Pei2020Geom-GCN:} and there are three subdatasets, Cornell, Texas, and Wisconsin. For each dataset, a web page is a node and a hyperlink is an edge. The bag-of-words representation of each web page is used as node features. We aim to classify the nodes into one of the five classes, student, project, course, staff, and faculty.

% \textbf{Wikipedia Network} \cite{musae}: Chameleon, squirrel, and crocodile are three topics of Wikipedia page-page networks. Nodes are articles from the English Wikipedia, and edges represent links between them. Node features indicate the occurrences of specific nouns in the articles. Based on the average monthly traffic of the web page, the nodes are divided into five classes.

% \textbf{Citation Networks} \cite{sen2008collective}: Cora, Citeseer, PubMed are homophilous citation network datasets. Cora consists of seven classes of machine learning papers. CiteSeer has six classes. Papers are represented by nodes, while citations between two papers are represented by edges. Each node has features defined by the words that appear in the paper's abstract.

\subsubsection{Implementation Details} %5.1
\label{sec:imple}
The performance on heterophilous datasets is evaluated on ten train/validation/test splits. We use five repeated experiments with different random seeds to evaluate the performance on homophilous datasets. We use the fixed splits provided by PyG \cite{Fey/Lenssen/2019} if available. Otherwise, we conduct experiments on ten random splits. For HP-GMN, all MLPs are implemented with two layers. We vary the number of memory unit $K$ among \{20, 50, 100, 200, 300, 500\} and the hidden number of each memory unit $hidden$ among \{100, 200, 300, 500\}. We add the Frobenius norm of the memory $||\mathbf{M}||_{\mathrm{F}}^2$ into the loss to prevent overfitting. Coefficients $\alpha$ and $\beta$ are set by grid search in \{0.0001, 0.001, 0.01, 0.1, 1, 10, 100\}.

\subsection{Node Classification on Graphs with Hetreophily} %5.2
To answer \textbf{RQ1}, we conduct node classification on heterophilous graphs and compare HP-GMN with the baselines. The average accuracy with standard deviation are shown as Table \ref{tab:results} and we have the following observations:
\begin{itemize}[leftmargin=*]
    \item GCN is a little better than MLP or even no better than MLP on heterophilous datasets because GCN fails to exploit the structure information of heterophilous graphs. Our HP-GMN considers heterophily properties, and the local statistics can capture heterophilous structures well; thus, it achieves much better performance than MLP and GCN.
    \item While methods employing higher-order neighborhood information improve the performance, the extent is limited because they aggregate all the higher-order neighbors regardless the homophily or heterophily. HP-GMN only aggregates from similar global patterns, which leads to further improvements in accuracy.
    \item Methods designed for heterophily have better performance than general graph methods even without global patterns. However, HP-GMN takes advantage of global patterns and outperforms all other baselines significantly.
\end{itemize}

% On WebKB datasets, MLP outperforms most GNNs because heterophilous structures provide harmful information. Though some methods using higher-order neighbors like MixHop and APPNP improve the performance of GCN, they are still not better than MLP. On Wikipedia datasets, GNNs giving better results than MLP reveals that graph structures can favor the prediction. But the improvement is limited, which means information propagation is still underexploited. Our HP-GMN outperforms all the benchmarks and significantly improves the prediction accuracy on heterophilous datasets. That is because HP-GMN learns the local and global information that are helpful for the prediction effectively.

\begin{table}[t!]
\captionsetup{font={scriptsize}}
\caption{Node classiﬁcation performance on homophily. (Accuracy(\%) ± Std.)}
\vspace{-1em}
\begin{center}
\begin{tabularx}{0.8\linewidth}{lCCC}
\toprule
Methods   &\textbf{Cora}     &\textbf{Citeseer}      &\textbf{Pubmed}    \\ 
\midrule
MLP                     &53.5±1.5         &51.9±1.5              &69.9±0.6          \\
GCN                     &78.5±1.4         &67.8±1.0              &76.6±1.1          \\
MixHop                  &77.0±0.7         &63.9±0.4              &74.3±0.5          \\
GCN+JK                  &76.7±2.4         &64.8±1.4              &75.0±0.8          \\
APPNP                   &83.1±0.5         &70.8±0.9              &79.9±0.3          \\
$\text{HP-GMN}_{\text{GCN}}$   &80.3±1.1         &69.0±0.9              &77.8±0.7            \\
$\text{HP-GMN}_{\text{APPNP}}$ &\textbf{84.7±0.6}         &\textbf{73.0±1.0}              &\textbf{82.4±1.3}           \\
\bottomrule
\end{tabularx}
\end{center}
\label{tab:homoresults}
\vskip -2em
\end{table}

\begin{figure}[h!]
\centering
\vskip -1.25em
\captionsetup{font={scriptsize}}
\begin{subfigure}{0.43\columnwidth}
    \centering
    \includegraphics[width=1\linewidth]{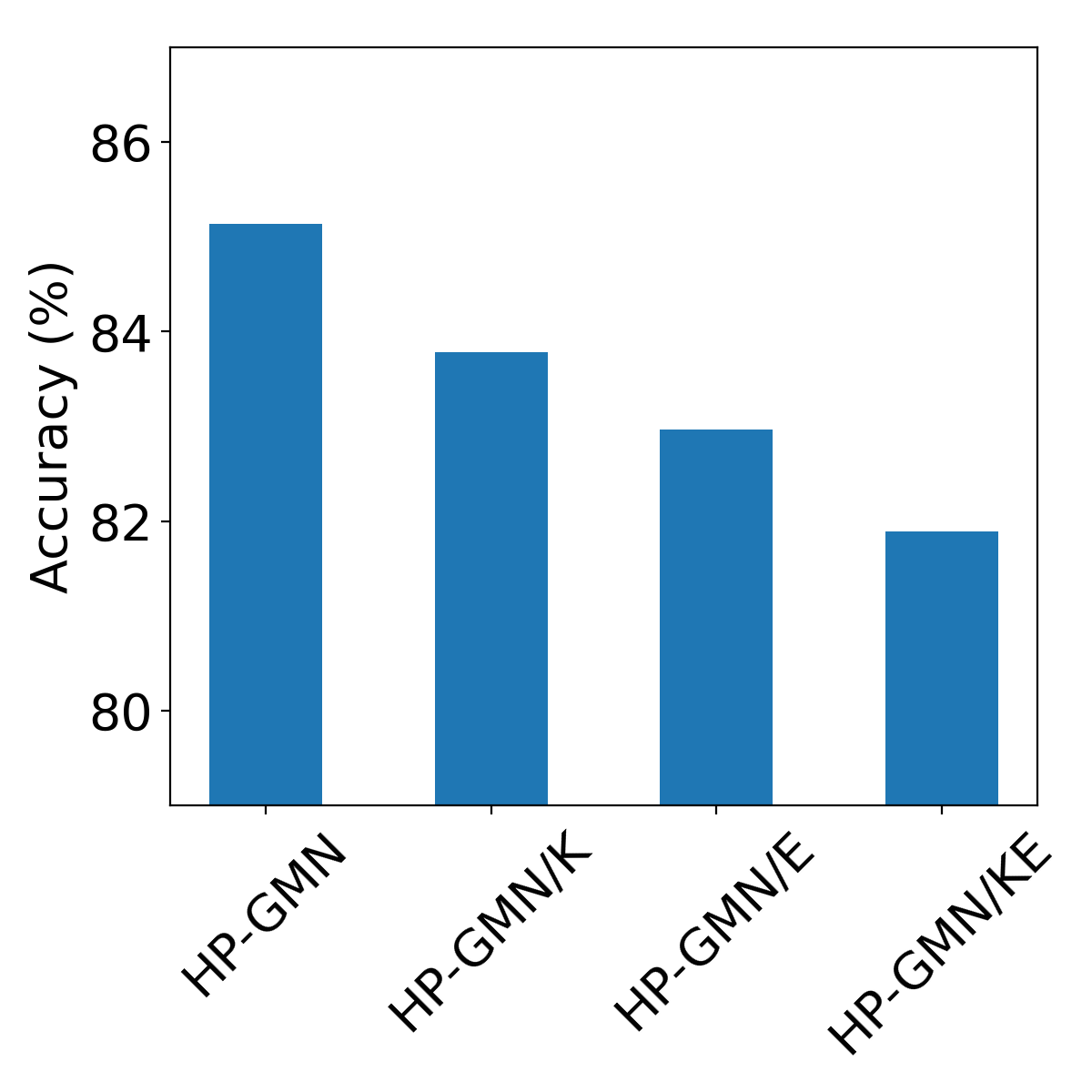} 
    \vskip -0.6em
    \caption{\scriptsize Texas}
    \label{fig:regu_texas}
\end{subfigure}
\begin{subfigure}{0.43\columnwidth}
    \centering
    \includegraphics[width=1\linewidth]{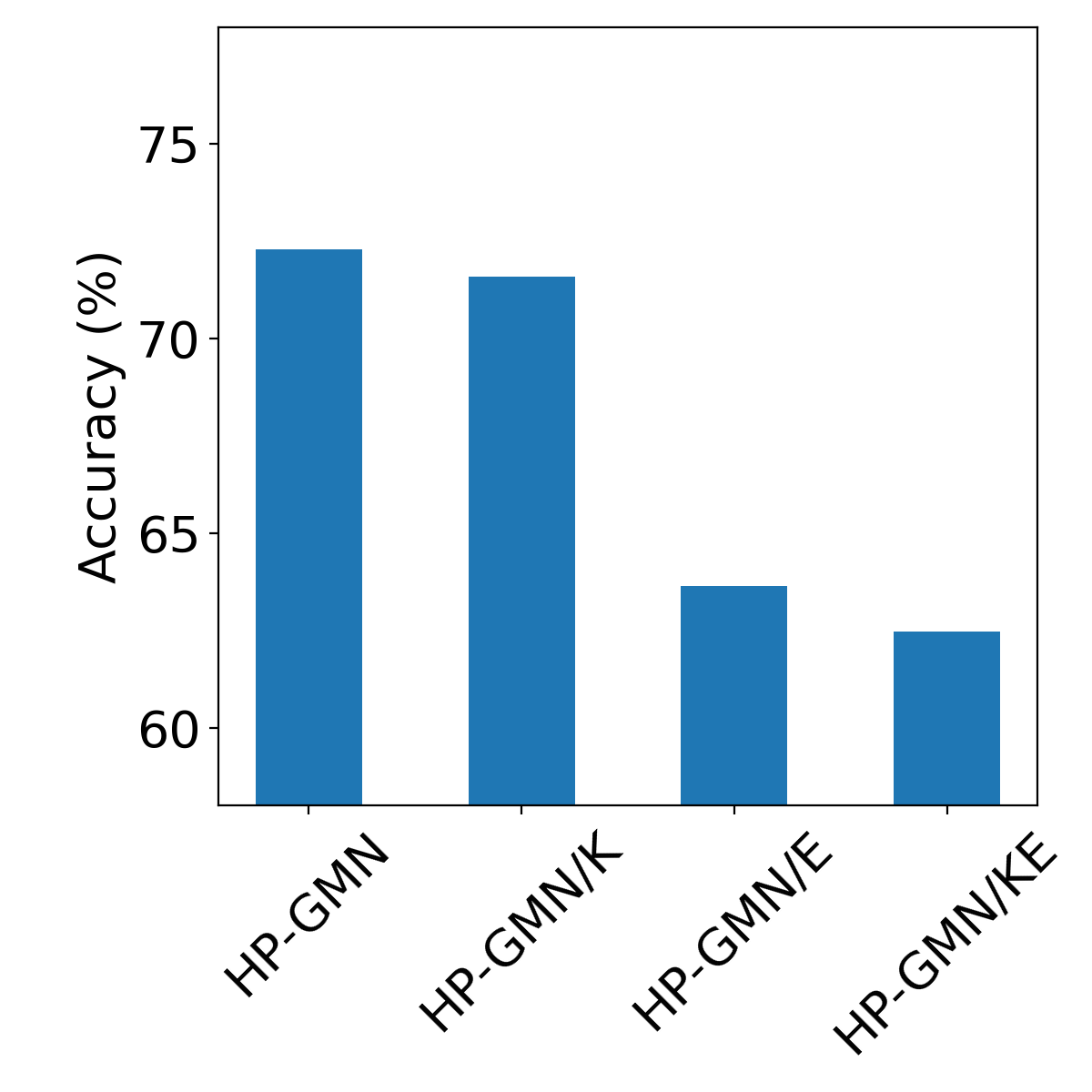} 
    \vskip -0.6em
    \caption{\scriptsize Squirrel}
    \label{fig:regu_squirrel}
\end{subfigure}
\vskip -0.3em
\caption{Impact of regularization terms on accuracy.}
\vskip -1.2em
\label{fig:regu}
\end{figure}

\begin{table}[h!]
\captionsetup{font={scriptsize}}
\caption{Impacts of local statistics.}
\vspace{-1em}
\begin{center}
\begin{tabularx}{0.8\linewidth}{p{0.35\linewidth}|CC}
\toprule
\multicolumn{1}{c|}{\textbf{Methods}}                &\multicolumn{1}{c}{\textbf{Texas}}         &\multicolumn{1}{c}{\textbf{Squirrel}}      \\
\midrule
All Local Statistics                                         &\textbf{85.1±4.2}     &\textbf{72.3±2.3}     \\ 
W/O Node Attributes                &70.3±5.5              &71.7±1.9              \\
W/O Label-wise Class        &83.2±6.0              &57.8±2.5              \\
W/O Label-wise Feature       &79.2±4.0              &41.7±3.4              \\
W/O Diffusion Matrix          &83.2±5.9              &68.4±6.6              \\
\bottomrule
\end{tabularx}
\end{center}
\vskip -2em
\label{tab:ablation_localstatistics}
\end{table}

\subsection{Node Classification on Graphs with Homophily} %5.3
To answer \textbf{RQ2}, we aim to demonstrate that HP-GMN can also work on homophilous datasets. However, the local statistics in \ref{local_stat} are designed for heterophily and do not fit the homophily problem. In order to show the effectiveness on the homophilous datasets, we use the representation learned by GCN or APPNP as the local statistic on graphs with homophily, denoted as $\text{HP-GMN}_{\text{GCN}}$ and $\text{HP-GMN}_{\text{APPNP}}$. We conduct experiments on three homophilous citation networks. The average accuracy with the standard deviation is shown in Table~\ref{tab:homoresults}. From Table~\ref{tab:homoresults}, we observe that the memory not only promotes the prediction on heterophily but also can facilitate learning on homophily. That is because homophilous graphs also contain some global information that is not captured by other methods.

\subsection{Ablation Study} %5.3
To answer \textbf{RQ3}, we compare the performance of HP-GMN, HP-GMN without Kpattern regularization (HP-GMN/K), HP-GMN without Entropy regularization (HP-GMN/E), and HP-GMN without regularizations (HP-GMN/KE). We report the performance on Texas and Squirrel in Fig.~\ref{fig:regu}. Then we show the contribution of local statistics. We remove each local statistic to get different designs of the query. We observe the best performance when we use all four local statistics. It reveals that all of them can collect different aspects of local information in the graph, and everyone is essential for describing the local subgraph. Results are shown in Table \ref{tab:ablation_localstatistics}, where ``W/O Node Attributes" denotes that ``Node Attributes" are removed from local statistics.

\section{Related Work} %5
Some methods for the heterophily modify the current GNN framework. $\text{H}_2\text{GCN}$ \cite{zhu2020beyond} uses ego and neighbor separation to encode the ego and neighbor representations separately instead of mingling them together. CPGNN \cite{zhu2021graph} uses the compatibility matrix to initialize and guide the propagation of the GNN. GPRGNN \cite{chien2021adaptive} combines intermediate representations and learn adaptive weights for them using Generalized PageRank. Some methods try to exploit global information by using higher-order neighborhoods. MixHop \cite{abu2019mixhop} learns higher-order information by utilizing multiple powers of the adjacency matrix. GCN-Cheby \cite{defferrard2016convolutional} uses k-order Chebyshev Polynomials to replace the first order Chebyshev Polynomials in GCN so that it can learn from up to k-order neighbors. Recently, \cite{xiao2022decoupled} also studied the problem of conducting self-supervised learning for node representation on heterophilous graphs when task-specific labels are scarce.

\section{Conclusion} %6
In this paper, we develop a novel graph memory network to address the heterophily problem. Local statistics and memory are utilized to capture local and global information in a graph. Kpattern and Entropy regularizations are proposed to encourage the memory to learn enough global information. Extensive experiments show that our method can achieve state-of-the-art performance on heterophilous and homophilous graphs.

% There are several future directions to be explored. First, better designs of local statistics and memory regularizations can prompt local and global information to be captured more efficiently. Moreover, all of the current works focus on node classification on heterophilous graphs. How to extend the graph memory network on some other downstream tasks, such as link prediction and graph clustering, are to be considered. 

\section*{Acknowledgment}
\vskip -0.5em
This project was partially supported by NSF projects IIS-1707548, CBET-1638320,  IIS-1909702 and IIS-1955851, the Army Research
Office under grant W911NF21-1-0198, and DHS CINA under grant E205949D.

\vskip -1.5em
\bibliography{ref}
\bibliographystyle{IEEEtran}

\end{document}